\documentclass[twocolumn,pre,floatfix,superscriptaddress]{revtex4}
\usepackage[T1]{fontenc}
\usepackage[latin9]{inputenc}
\usepackage{mathrsfs}
\usepackage{amsmath}
\usepackage{amssymb}
\usepackage{graphicx}
\usepackage{braket}
\usepackage{color}
\usepackage[svgnames]{xcolor}
\usepackage{comment}
\usepackage{natbib}
\usepackage{dsfont}
\usepackage{mdframed}

\makeatletter

\usepackage{listings}

\usepackage{hyperref}

\hypersetup{
  colorlinks=true,
  linkcolor=blue,
  citecolor=ForestGreen,
  urlcolor=DarkOrchid
}

\AtBeginDocument{
  
}

\makeatother

\usepackage{babel}

\begin{document}

\title{Reply to: Inability of a graph neural network heuristic to outperform greedy algorithms in solving combinatorial optimization problems}

\author{Martin~J.~A.~Schuetz}
\affiliation{Amazon Quantum Solutions Lab, Seattle, Washington 98170, USA}
\affiliation{AWS Center for Quantum Computing, Pasadena, CA 91125, USA}

\author{J.~Kyle~Brubaker}
\affiliation{Amazon Quantum Solutions Lab, Seattle, Washington 98170, USA}

\author{Helmut G.~Katzgraber}
\affiliation{Amazon Quantum Solutions Lab, Seattle, Washington 98170, USA}
\affiliation{AWS Center for Quantum Computing, Pasadena, CA 91125, USA}

\date{\today}

\begin{abstract}

We provide a comprehensive reply to the comment written by Stefan Boettcher [arXiv:2210.00623] and argue that the comment singles out one particular non-representative example problem, entirely focusing on the maximum cut problem (MaxCut) on \textit{sparse} graphs, for which greedy algorithms are expected to perform well. 
Conversely, we highlight the broader algorithmic development underlying our original work \citep{schuetz:22}, and (within our original framework) provide additional numerical results showing sizable improvements over our original data, thereby refuting the comment's original performance statements. 
Furthermore, it has already been shown that physics-inspired graph neural networks (PI-GNNs) can outperform greedy algorithms, in particular on hard, \textit{dense} instances. 
We also argue that the internal (parallel) anatomy of graph neural networks is very different from the (sequential) nature of greedy algorithms, and (based on their usage at the scale of real-world social networks) point out that graph neural networks have demonstrated their potential for superior scalability compared to existing heuristics such as extremal optimization. Finally, we conclude highlighting the conceptual novelty of our work and outline some potential extensions. 

\end{abstract}

\date{\today}

\maketitle


\textbf{Problem instances and benchmarks.} 
The comment exclusively focuses on results for {\em{sparse}} random $d$-regular graphs and does not account for results on the publicly-available \texttt{Gset} benchmark instances for which we list results in Table I of our paper \citep{schuetz:22}. While the comment claims that there is a lack of state-of-the-art comparisons, here we do report on a wide array of benchmark results, including results based on an SDP solver using dual scaling (DSDP), Breakout Local Search (BLS), a Tabu Search metaheuristic (KHLWG), and a recurrent GNN architecture for maximum constraint satisfaction problems (RUN-CSP). We find that our simple graph convolutional network (GCN) baseline architecture is competitive with these solvers and typically within approximately $1\%$ of the best results based on BLS. In addition, we have chosen to provide results for random $d$-regular results primarily because these allow us to perform large-scale experiments and averaging over instances, as well as comparisons to analytical bounds. At large scales these instances are sparse, and it is not surprising that greedy algorithms perform well in this case. We do not think that greedy algorithms (while performant for sparse instances) provide a universal baseline, and still think that the Goemans-Williamson algorithm was a reasonable algorithmic choice to compare to, as it is a fairly established, widely-used algorithm for which we have shown on-par performance, with much better scalability. In fact, we have highlighted the shortcomings of greedy algorithms in our follow-up work on graph coloring problems where we have also presented results demonstrating that physics-inspired graph neural networks (PI-GNNs) can outperform greedy algorithms, in particular on \textit{dense} instances \citep{schuetz:22b}.

\textbf{Algorithmic development.} 
The comment seemingly equates the GNN heuristic with the vanilla graph convolutional network (GCN) layer architecture we initially used for demonstration purposes. However, graph neural networks represent an entire growing list of neural networks under the larger umbrella of geometric deep learning \citep{bronstein:21}. Thus, there is not just one GNN architecture just as there is not just one genetic algorithm, but rather a whole family of GNN-based architectures within one larger framework. In our original numerical experiments we have restricted ourselves to a simple two-layer GCN architecture as it represents a simple baseline model. Indeed, we could have used more complex GNN architectures as shown below. However, we wanted to compare against a simple baseline model. As such, the conceptual novelty is much broader.

\begin{figure}
\includegraphics[width=1.0 \columnwidth]{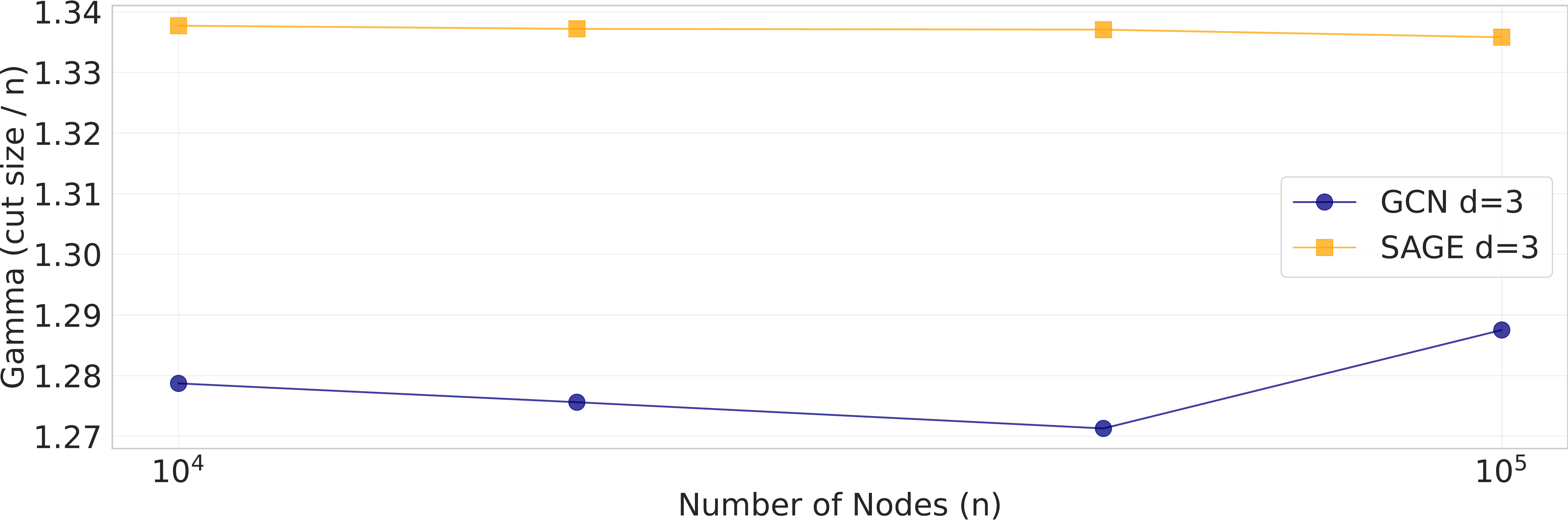}
\caption{
Relative cut size $(\mathrm{cut}/n)$ for 3-regular graphs, as a function of the number of nodes $n$. In addition to the GCN-based results (dark circles, as previously reported) numerical (average) results are shown for a GraphSAGE architecture (light squares). GraphSAGE consistently provides larger cut values, roughly by a factor of $\sim 1.337/1.28 \approx 1.0445$. All data points are averaged results, with $20$ samples per problem size. 
\label{fig:cut_sage}}
\end{figure}

To demonstrate this statement, we have performed additional numerical experiments for MaxCut on 3-regular graphs with $n \in [10^4,10^5]$. The results are shown in Fig.~\ref{fig:cut_sage}. By simply replacing the GCN module with GraphSAGE \citep{hamilton:17} (which amounts to changing just a few lines of code) we observe a sizable improvement in the average cut size. Specifically, we find $\gamma_{3} \sim 1.337$ for GraphSAGE, as opposed to $\gamma_{3} \sim 1.28$ as previously reported for the vanilla GCN baseline. Converting to the proposed figure of merit (with $\left<e\right>$ denoting the average ground-state energy density) this translates into an improvement of $\sim 10.78\%$ (i.e., much closer to the optimum, and clearly better than the greedy baseline), because $\left< e_{3}\right>/\sqrt{3} \approx -0.678$ for GraphSAGE, while $\left< e_{3}\right>/\sqrt{3} \approx -0.612$ for the GCN-based architecture. In a second iteration of the comment Boettcher then presented results for an extended greedy local search that performs slightly better (approximately $\sim 2\%$) than our GraphSAGE results on these selected, sparse instances.   Again, we are not surprised that such a heuristic works well for \textit{sparse} instances \citep{pang:21}, and refer to Ref.~\citep{schuetz:22b} showing that our general-purpose heuristic dubbed PI-SAGE can outperform greedy heuristics on \textit{dense} instances. Below we outline more approaches for further, potential improvements compared to the results presented here. Finally, we point out that conceptually the comparison to greedy algorithms is somewhat misleading. In fact, the internal anatomy of PI-GNN is very different from greedy algorithms, because updates to node representations are done in a fully parallelized fashion, as opposed to the sequential nature of greedy methods. 

\textbf{Presentation of data.} 
The comment argues for a particular presentation of the data, using an inverted $x$-axis $(\sim 1/n)$; this approach is certainly interesting, but comes with its own downsides. 
Specifically, all large-scale results are squashed together close to the $y$-axis, with no discernible difference between $n \sim 10^4$ and $n \sim 10^6$, thereby hiding the limited scalability of some existing solvers, while not accurately presenting the scalability of the proposed PI-GNN solver. 
In particular, we disagree with the scalability-related comment (``\textit{... likely a significant overhead in the pre-factor due to the GNN itself}''). We have shown that PI-GNNs can solve MaxCut problems with $n \sim 10^6$ variables in approximately 15 minutes, as can be easily seen from the simple linear x-axis $\sim n$ of our figures. Moreover, using distributed training, GNNs have been trained on billion-scale instances \citep{zheng:20}, for example in the context of social networks, thus demonstrating that there is no significant overhead as stated in the comment. The proposed form of data presentation may be helpful in the evaluation of asymptotic scaling for an extremely narrow set of benchmark instances where configurational averages can be performed, and that seems to be the comment's prime motivation. However, this was not our intention in Ref.~\citep{schuetz:22}. Depending on the context, extrapolated asymptotics can be of interest, but in practice we see many industry use cases (such as, for example, portfolio optimization in finance) where customers need actual, constructive solutions (like a bitstring $\textbf{x}$) for problems with more than $10^5$ variables often within a limited time budget. Note also that the extrapolated asymptotic limit of the energy is of little practical value in such a setting. Finally, we would like to point out that many heuristics such as extremal optimization require sizable computational time to address problems with $10^5$ or more variables, thus rendering these ineffective in that regime.

\textbf{Conceptual novelty.} 
The comment is limited in scope and does not pay much attention to the conceptual novelty of our work. It is focused on the MaxCut problem, whereas we outline broader applicability to other problems, within the large family of QUBO and PUBO problems. In addition, we recently showed that the PI-GNN framework can be readily extended to entirely new problem classes, such as the graph coloring problem \citep{schuetz:22b}. As mentioned in our original paper \citep{schuetz:22}, we believe that our work can motivate and trigger a vast array of interesting follow-up studies, including (but not limited to) more detailed analysis of the limitations of the proposed GNN-based framework. For example, we have already shown how simple tweaks of the GNN architecture can result in sizeable performance improvements \citep{schuetz:22b}. Apart from GraphSAGE, alternative candidates are Graph Attention Networks (GATs) or Graph Isomorphism Networks (GINs) \citep{xu:19}, among others. In addition, beyond our vanilla baseline approach, more sophisticated training schemes (for example, incorporating ideas from annealing), graph rewiring strategies to decouple the GNN training graph from the problem graph \citep{topping:21}, schemes to incorporate structural graph features into the node embeddings, as well as more refined post-processing techniques should all help improve the performance of the proposed GNN approach. We feel that the myriad of these exciting extensions of our research are not reflected in this comment.

{\bf Competing interests.} M.J.A.S., J.K.B. and H.G.K. are listed as inventors on a US provisional patent application (no. 7924-38500) on combinatorial optimization with graph neural networks.

\textbf{Correspondence and requests for materials} should be addressed to Martin J. A. Schuetz, J. Kyle Brubaker or Helmut G. Katzgraber.

\bibliography{refs}

\end{document}